\documentclass[letterpaper, 10 pt, conference]{ieeeconf}

\IEEEoverridecommandlockouts

\usepackage{amsmath,amsfonts}
\usepackage{algpseudocode}
\usepackage{algorithm}
\usepackage{array}
\usepackage{textcomp}
\usepackage{stfloats}
\usepackage{url}
\usepackage{verbatim}

\usepackage[utf8]{inputenc}
\usepackage[T1]{fontenc}
\usepackage{graphicx}
\usepackage{caption}
\usepackage{subcaption}
\usepackage{makecell}

\usepackage{graphics} 
\usepackage{epsfig} 
\usepackage{xcolor}
\usepackage[normalem]{ulem}
\usepackage{booktabs}
\usepackage{multicol}
\usepackage{multirow}
\usepackage{cancel}
\usepackage{cite}
\usepackage{cleveref}
\usepackage{comment}
\usepackage{siunitx}
\let\labelindent\relax
\usepackage[inline]{enumitem}

\usepackage{acro}
\usepackage{bm}
\usepackage{mathtools}
\usepackage{xcolor}

\DeclareAcronym{ASL}{short = ASL, long = Autonomous Systems Lab}
\DeclareAcronym{OMAV}{short = OMAV, long = Omnidirectional Micro Aerial Vehicle}
\DeclareAcronym{MAV}{short = MAV, long = Micro Aerial Vehicle}
\DeclareAcronym{UAV}{short = UAV, long = Unmanned Aerial Vehicle}
\DeclareAcronym{RC}{short = RC, long = Remote Controller}
\DeclareAcronym{DoF}{short = DoF, long = Degrees of Freedom}
\DeclareAcronym{PBC}{short = PBC, long = passivity-based control}
\DeclareAcronym{PH}{short = PH, long = Port-Hamiltonian}
\DeclareAcronym{NDT}{short = NDT, long = non-destructive testing}
\DeclareAcronym{PEMS}{short = PEMS, long = Power and Energy Monitoring System}
\DeclareAcronym{WTC}{short = WTC, long = wrench tracking controller}
\DeclareAcronym{PTC}{short = PTC, long = pose tracking controller}
\DeclareAcronym{MBE}{short = MBE, long = momentum-based wrench estimator}
\DeclareAcronym{ASIC}{short = ASIC, long = Axis-Selective Impedance Control}
\DeclareAcronym{MPC}{short = MPC, long = Model Predictive Control}
\DeclareAcronym{MPPI}{short = MPPI, long = Model Predictive Path Integral}
\DeclareAcronym{APhI}{short = APhI, long = Aerial Physical Interaction}
\DeclareAcronym{LLE}{short = LLE, long = Largest Lyapunov Exponent}
\DeclareAcronym{ICBF}{short = ICBF, long = Integral Control Barrier Function}
\DeclareAcronym{CBF}{short = CBF, long = Control Barrier Function}
\DeclareAcronym{COM}{short = CoM, long = Center of Mass}
\DeclareAcronym{AM}{short = AM, long = Aerial Manipulator}
\DeclareAcronym{MR}{short = MR, long = Mixed Reality}
\DeclareAcronym{AR}{short = AR, long = Augmented Reality}
\DeclareAcronym{VR}{short = VR, long = Virtual Reality}
\DeclareAcronym{HRI}{short = HRI, long = Human-Robot Interaction}
\DeclareAcronym{RL}{short = RL, long = Reinforcement Learning}
\DeclareAcronym{PPO}{short = PPO, long = Proximal Policy Optimization}

\DeclareAcronym{NASA-TLX}{short = NASA-TLX, long = NASA Task Load Index}
\DeclareAcronym{MD}{short = MD, long = mental demand}
\DeclareAcronym{PD}{short = PD, long = physical demand}
\DeclareAcronym{TD}{short = TD, long = temporal demand}
\DeclareAcronym{EF}{short = EF, long = effort}
\DeclareAcronym{PE}{short = PE, long = performance}
\DeclareAcronym{FR}{short = FR, long = frustration}
\DeclareAcronym{SNR}{short = SNR, long = signal-to-noise ratio}
\DeclareAcronym{ANOVA}{short = ANOVA, long = Analyse of Variance}
\DeclareAcronym{FT}{short = F/T, long = force and torque, short-indefinite = an, long-indefinite = a}
\DeclareAcronym{BBT}{short = BBT, long = Box and Block Test}
\DeclareAcronym{ABBT}{short = ABBT, long = Aerial Box and Block Test}
\DeclareAcronym{MOCAP}{short = MOCAP, long = Motion Tracking System}

\DeclareAcronym{FC}{short = FC, long = Flight Controller}
\DeclareAcronym{SITL}{short = SITL, long = Software In The Loop}
\DeclareAcronym{IMU}{short = IMU, long = Inertial Measurement Unit}

\renewcommand{\vec}[1]{\bm{#1}}		
\newcommand{\matr}[1]{\bm{#1}}		
\newcommand{\nR}[1]{\mathbb{R}^{#1}}		
\newcommand{\SO}[1]{\mathsf{SO}(#1)}		
\newcommand{\upperRomannumeral}[1]{\uppercase\expandafter{\romannumeral#1}}	

\newcommand{\transpose}{^\top}

\renewcommand{\frame}[1]{\mathcal{F}_{#1}}		

\newcommand{\origin}{O}						
\newcommand{\vX}{\vec{x}}					

\newcommand{\vY}{\vec{y}}					
\newcommand{\vZ}{\vec{z}}					
\newcommand{\pos}{\vec{p}_B}				
\newcommand{\posRef}{\vec{p}_{B,\text{ref}}}    
\newcommand{\vel}{\vec{v}_B}				
\newcommand{\velRef}{\vec{v}_{B,\text{ref}}}	
\newcommand{\accB}{\dot{\vec{v}}_B^B}	

\newcommand{\velMax}{{v}_{max}}
\newcommand{\rotMat}{\matr{R}}				

\newcommand{\frameW}{\frame{W}}			
\newcommand{\frameB}{\frame{B}}			
\newcommand{\frameM}{\frame{M}}			
\newcommand{\frameC}{\frame{C}}			
\newcommand{\frameR}{\frame{R}}			
\newcommand{\frameL}{\frame{L}}			
\newcommand{\originW}{\origin_W}		
\newcommand{\originB}{\origin_B}		
\newcommand{\originC}{\origin_C}		
\newcommand{\originR}{\origin_R}		
\newcommand{\originL}{\origin_L}		
\newcommand{\xW}{\vX_W}				
\newcommand{\yW}{\vY_W}				
\newcommand{\zW}{\vZ_W}				
\newcommand{\xB}{\vX_B}				
\newcommand{\yB}{\vY_B}				
\newcommand{\zB}{\vZ_B}				
\newcommand{\xM}{\vX_M}				
\newcommand{\zM}{\vZ_M}				
\newcommand{\xC}{\vX_C}				
\newcommand{\yC}{\vY_C}				
\newcommand{\zC}{\vZ_C}				
\newcommand{\xR}{\vX_R}				
\newcommand{\yR}{\vY_R}				
\newcommand{\zR}{\vZ_R}				
\newcommand{\xL}{\vX_L}				
\newcommand{\yL}{\vY_L}				
\newcommand{\zL}{\vZ_L}				

\newcommand{\rotMatWB}{\rotMat_B^W}	
\newcommand{\rotMatWBRef}{\rotMat_{B,\text{ref}}^W}	
\newcommand{\rotMatL}{\rotMat_{L}}	
\newcommand{\rotMatR}{\rotMat_{R}}	
\newcommand{\rotMatC}{\rotMat_{C}}	
\newcommand{\rotMatCL}{\rotMat_L^C}	
\newcommand{\rotMatCR}{\rotMat_R^C}	
\newcommand{\rotMatWC}{\rotMat_C^W}	

\newcommand{\angVel}{\vec{\omega}_B}
\newcommand{\angRotR}{\vec{\omega}_R}
\newcommand{\angRotL}{\vec{\omega}_L}

\newcommand{\angVelR}{\dot{{\vec{\omega}}}_R}
\newcommand{\angVelL}{\dot{{\vec{\omega}}}_L}
\newcommand{\angVelB}{\dot{{\vec{\omega}}}_B^B}

\newcommand{\wrenchExt}{\wrench_\text{ext}}
\newcommand{\wrenchEst}{\hat{\wrench}_\text{ext}}
\newcommand{\wrenchTotal}{\wrench_\text{fb,total}}
\newcommand{\wrenchRec}{\wrench_\text{fb,rec}}
\newcommand{\wrenchFbExt}{\wrench_\text{fb,ext}}

\newcommand{\wrenchLAct}{\wrench_\text{L,act}}
\newcommand{\wrenchL}{\wrench_\text{L}}
\newcommand{\wrenchRAct}{\wrench_\text{R,act}}
\newcommand{\wrenchR}{\wrench_\text{R}}

\newcommand{\wrench}{\bm{\tau}}

\title{\LARGE \bf
Design and Control of an Omnidirectional Aerial Robot\\with a Miniaturized Haptic Joystick for Physical Interaction
}

\author{Julien Mellet$^{*1}$, Andrea Berra$^{*2}$, Salvatore Marcellini, Miguel Ángel Trujillo Soto$^{2}$, Guillermo Heredia$^{3}$,\\ Fabio Ruggiero$^{1}$, Vincenzo Lippiello$^{1}$
\thanks{This project has received funding from the European Union’s Horizon 2020 research and innovation program under the Marie Skłodowska-Curie grant agreement No 953454.}
\thanks{$^{1}$PRISMA Lab, Department of Electrical Engineering and Information Technology, University of Naples Federico II Naples, Italy.}
\thanks{$^{2}$CATEC, Advanced Center for Aerospace Technologies, Seville, Spain.}
\thanks{$^{3}$GRVC, Robotics, Vision and Control Group
School of Engineering, University of Seville
Seville, Spain.}
\thanks{$^{*}$The authors contributed equally}
\thanks{Corresponding Authors' email: julien.mellet@unina.it}
}

\begin{document}

\maketitle
\thispagestyle{empty}
\pagestyle{empty}

\begin{abstract}
Fully actuated aerial robot proved their superiority for \ac{APhI} over the past years. This work proposes a minimal setup for aerial telemanipulation, enhancing accessibility of these technologies. The design and the control of a 6\mbox{-}\ac{DoF} joystick with 4\mbox{-}\ac{DoF} haptic feedback is detailed. It is the first haptic device with standard \ac{RC} form factor for \ac{APhI}. By miniaturizing haptic device, it enhances \ac{RC} with the sense of touch, increasing physical awareness. The goal is to give operators an extra sense, other than vision and sound, to help to perform safe \ac{APhI}. To the best of the authors knowledge, this is the first teleoperation system able to decouple each single axis input command. On the omnidirectional quadrotor, by reducing the number of components with a new design, we aim a simplified maintenance, and improved force and thrust to weight ratio. Open-sourced physic based simulation and successful preliminary flight tests highlighted the tool as promising for future \ac{APhI} applications.
\end{abstract}

\begin{figure*}[hb]
  \centering
  \includegraphics[width=\linewidth]{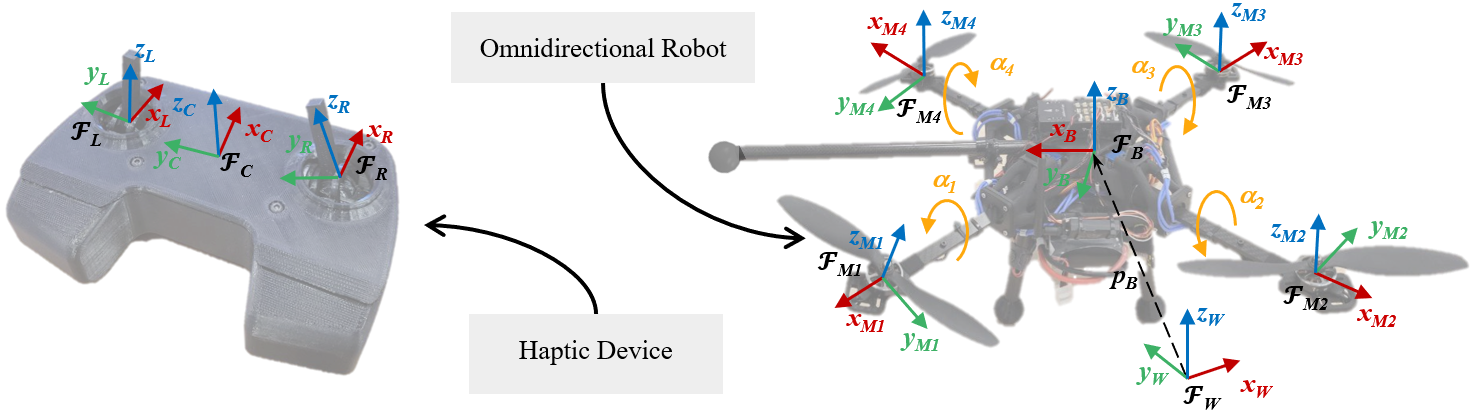}
  \caption{The telemanipulation system includes the joystick (\textit{left}) and the aerial robot (\textit{right}). The haptic joystick with its frame $\frameC$ and miniaturized force feedback mechanism on the sticks with respective left $\frameL$ and right $\frameR$ frames. The omnidirectional quadrotor with its body frame $\frameB$, in position $\pos$ with respect to the world frame $\frameW$. The four tilting rotor frames $\frameM$ are with $\zM$ pointing up and $\xM$ along the robot arms with $\alpha$ angles around them.}
  \label{fig:eye-catcher}
\end{figure*}

\section{Introduction}
\label{sec:introduction}

%






The field of \ac{UAV}s has evolved from aeromodelism technologies, leading to the usage of 4\mbox{-}\ac{DoF} radio controllers for piloting the majority of the multirotor platform. Similarly, ground robots are often operated using gamepad-like controllers. However, with the advent of \ac{OMAV}s~\cite{past-future-am}, which offer full actuation capabilities for aerial physical interaction (APhI), limitations in current control interfaces are becoming evident. Specifically, there is a shortage of input channels and a lack of tactile feedback to inform the operator during complex maneuvers. Haptic Devices, mainly developed for medical purposes~\cite{selvaggio2021haptics}, could address these limitations. Within the context of aerial robotics, integrating force and tactile feedback into conventional controller designs, operators can be better informed about external forces acting on the vehicle, such as wind disturbances or physical interactions with surfaces.  Furthermore, optimizing the interface for seamless integration with full-actuation platforms, such as \ac{OMAV}s, would allow operators to fully exploit the agility and control potential of these vehicles, leading to more stable and responsive flight performance.

This paper seeks to bridge the gap between research on haptic aerial control and industrial inspection by miniaturizing the interface into a conventional joystick device. It presents an open-source telemanipulation system that can be replicated using standardized tools within a unified teleoperation framework. To the best of the author's knowledge, this work is the first to integrate standard \ac{RC} joysticks with significant force feedback into a operational omnidirectional device.

\subsection{Related Works} 
Omnidirectional platforms are intricate machines with numerous components, requiring substantial maintenance due to the complexity of their systems and current manufacturing techniques. As a result, these platforms often have a low thrust-to-weight ratio~\cite{bodie2020active}. The integration of force sensors for contact inspection further increases their intricacy. Therefore, omnidirectional aerial robots designed for physical interactions are inherently sophisticated systems.

To facilitate their deployment and ensure safety, simulating omnidirectional aerial robots is essential for accelerating their development. OmniDrones~\cite{botian2024omni} is a flexible simulator for \ac{OMAV}s. It is oriented for freeflight reinforcement learning training but does not embed \ac{FC} \ac{SITL}.

For bilateral teleoperation~\cite{bilateral} of omnidirectional robots, hand-held joysticks are commonly employed~\cite{mike-6dof, rotation-ctl}. These joysticks cannot fully decouple rotational movements from unintended translations, severely limiting the applicability of 6-DoF systems. Additionally, a widespread use of these joystick would necessitate retraining of teleoperators. Although a 3-translational DoF haptic stick has been proposed~\cite{mintchev2019portable}, its lack of force intensity and shows fragility due to its foldable mechanism. Therefore origamic mechanism is unsuitable for industrial use because robust force feedback is essential for safe aerial vehicle steering.

Thus, there is a need for a portable, on-site joystick that keeps the current RC form factor while enhancing teleoperator expertise through haptic augmentation. However, the challenge of mapping 4-DoF inputs to the 6-DoF capabilities of the OMAV remains unresolved.



Even if \ac{APhI} applications have been achieved over the past year, no previous research managed to provide an entire aerial telemanipulation system for the community.

\subsection{Contributions}
This work proposes and shares a full system for aerial physical interaction in the haptic telemanipulation context including the robot with its joystick and simulator. The main contributions include
\begin{enumerate*}[label=\textit{\roman*)}]
    \item the simplification and open-sourcing of the design for an omnidirectional aerial robot, including its \ac{FC},
    \item the design of a miniaturized haptic joystick with standard \ac{RC} form factor, and 
    \item evaluation of the proposed telemanipulation system showing single axis command capability.
\end{enumerate*}
    
\section{Omnidirectional platform}\label{sec:platform}

\subsection{Model}
The omnidirectional multirotor in its environment is defined by the inertial world frame $\frameW$ (see Fig.~\ref{fig:eye-catcher}), where $\originW$ is the origin and ${\xW,\yW,\zW}$ are unit axes, located at an arbitrary fixed point such that $\zW$ points in the opposite direction of gravity. The aerial robot is described in the body frame $\frameB={\originB,\xB,\yB,\zB}$, where the origin $\originB$ coincides with the robot's \ac{COM}, and $\xB$ is oriented towards the end-effector stick. The system state uses $\pos \in \nR{3}$, the position of $\originB$ relative to $\frameW$. The attitude of $\frameB$ with respect to $\frameW$ is described by the rotation matrix $\rotMatWB \in \SO{3}$. From $\rotMatWB$, we can extract the Euler angles $(\phi_B,\theta_B,\psi_B)$ to express the robot's attitude in $\frameB$ with respect to $\frameW$. Also, the body linear and angular velocities with respect to $\frameW$ are denoted as $\vel$ and $\angVel \in \nR{3}$, respectively. 

The simplified system dynamics are derived in the Lagrangian form as in~\cite{bodie2020active, brunner2022planning}
\begin{equation}
\bm {M \dot{v} }+ \bm{C v} + \bm g = \bm{\tau_a} + \bm{\tau_e},
\end{equation}
where $\bm v$ and $\bm \dot{v} \in \mathbb{R}^{6 \times 1}$ are stacked linear and angular velocity, and linear and angular acceleration of the origin, $\bm M \in \mathbb{R}^{6 \times 6}$ is the symmetric positive definite inertia matrix, $\bm C \in \mathbb{R}^{6 \times 6}$ contains the centrifugal and Coriolis terms, and $\bm g \in \mathbb{R}^{6 \times 1}$ is the gravity vector. The external wrench acting on the platform is denoted by $\wrenchExt \in \nR{6}$, while the one related to the actuators is denoted by $\wrench_a \in \nR{6}$.

\subsection{Control}
For compliant interaction, the robot is controlled by an impedance controller~\cite{bodie2020active} in combination with a momentum-based external wrench estimator. This results in the following closed-loop dynamics
\begin{equation}
    \label{haptic_model}
    \Vec{M}_{v} \begin{bmatrix} \accB \\ \angVelB \end{bmatrix} + \Vec{D}_{v} \begin{bmatrix} \Vec{e}_v \\ \Vec{e}_{\omega} \end{bmatrix} + \Vec{K}_v \begin{bmatrix} \Vec{e}_p \\ \Vec{e}_{R} \end{bmatrix} = \wrench_a + \wrenchExt,
\end{equation}
where $\mathbf{M}_{v}$, $\mathbf{D}_{v}$, $\mathbf{K}_{v} \in \nR{6\times6}$ are the virtual inertia, damping, and stiffness matrices, respectively, which are given parameters. The position $\Vec{e}_p$, orientation $\Vec{e}_R$, velocity $\Vec{e}_v$ and angular rate $\Vec{e}_{\omega}$ errors as
\begin{subequations}
 	\label{position_errors}
 	\begin{IEEEeqnarray} {ll}
 		\Vec{e}_p &= \rotMatWB {\transpose} \left( \pos - \posRef \right),\\
         \Vec{e}_R &= \frac{1}{2}\left( \rotMatWBRef{\transpose} \rotMatWB - \rotMatWB {\transpose} \rotMatWBRef \right)^\vee,\\
         \Vec{e}_v &= \rotMatWB {\transpose} \left( \vel - \velRef \right),\\
         \Vec{e}_{\omega} &= \Vec{\omega}_B - \rotMatWB {\transpose}  \rotMatWBRef \Vec{\omega}_{B,ref},
 	\end{IEEEeqnarray}
\end{subequations}
with $\left(\cdot\right)^\vee$ the Vee operator to extract a vector from a skew-symmetric matrix.
The reference position and orientation, 
$\posRef$ and $\rotMatWBRef$, respectively, are computed later in~\eqref{eq:ref-input}.

\subsection{Control allocation}\label{sec:mixer}
The control allocation problem for an omnidirectional multirotor with tiltable arms consists of finding the optimal rotors' speeds $\omega_1, ..., \omega_4$ and, differently from standard multirotors, the optimal tilt-angles $\alpha_1, ..., \alpha_4$ to generate the desired wrench $\wrench_a$. This problem has been solved as in~\cite{Kamel2018}, where the allocation matrix is formulated to distinctly separate the vertical and lateral forces produced by each rotor. It is essential to account for uncertainties when utilizing cost-effective actuators, such as servomotors, for the actuation of the tiltable arms. Servomotors typically exhibit minimal positioning errors in steady-state conditions; however, when considering the inertia of the propulsion system (motor and propeller), they may experience non-negligible uncertainties in angular velocity, particularly during transient phases. Also, due to the critical coupling between the evolution of the tilt angles and the rotors' speed, even slight mismatches can lead to the generation of a wrench that differs from the intended one. Experiments have shown that this effect can lead to significant uncontrolled platform's behavior. The least significant effect arises from minor positioning errors during transient phases, which can be mitigated through appropriate controller tuning. In contrast, the most pronounced impact of these uncertainties is observed in the yaw angle dynamics, where small tilt angle deviations can induce oscillations and potentially lead to instability. To address this issue, a damping constant has been incorporated into the static allocation matrix $ \bm A_{s}$ proposed in~\cite{Kamel2018}, in order to reduce the required lateral force for generating the desired moments
\begin{subequations}
    \begin{align}
    A(\phi,i)_{s} &= k_{\phi} A^l(\psi,i)_{s} + (1-k_{\phi})A^v (\phi,i)_{s}, \\
    A(\theta,i)_{s} &= k_{\theta} A^l(\psi,i)_{s} + (1-k_{\theta})A^v (\theta,i)_{s}, \\
    A(\psi,i)_{s} &= k_{\psi} A^l(\psi,i)_{s} + (1-k_{\psi})A^v (\psi,i)_{s},
    \end{align}
    \label{eq:damping_constant}
\end{subequations}
where $(k_{\phi}, k_{\theta}, k_{\psi}) \in [0,1]$ represent the damping coefficients for roll, pitch, and yaw, respectively, while $(A^l(.)_s, A^v(.)_s)$ denote the lateral and vertical force contributions to the mixer. If $k_i > 0$, the corresponding moment is coupled with the rotors' tilt angles. As an experimental results shown in Fig.~\ref{fig:servo_rotor_flight}, even with servomotor uncertainties in angular speed, the robot demonstrated accurate tracking performance.

\subsection{Mechanical Design} 

The proposed design links the four landing gears with the four tilting arms as presented in Fig.~\ref{fig:eye-catcher}. The different parts of the robot and its assembly are shared on GitHub\footnote{https://github.com/tilties2/Haptic-OmniQuad.git}. Similar to the platform in~\cite{hui2024passive}, we decided to keep a unique rotor per robot arm in order to simplify previous design~\cite{bodie2020active} with unwanted aerodynamic effect of coaxial-rotor. Compared to omnidirectional hexarotor architecture, having only four arms reduces the maximum force and torque envelop in the plan $(\originB,\xB,\yB)$. Nevertheless, our architecture simplifies the overall robot maintenance. By minimizing the number of rotors and adopting $10in$ propellers, we achieve a compact wingspan, while maintaining a good overall efficiency.

The summary of the platform's main components and the corresponding weight is proposed in Table \ref{tab:drone_components}.
\begin{table}[ht]
    \centering
    \begin{tabular}{|l|l|l|}
    \hline
    \textbf{Component} & \textbf{Name} & \textbf{Weight [g]} \\
    \hline
     Servos & SPT4412LV & 260 \\
     Rotors & XING2 2809 1250KV & 243 \\
     Battery & 4s 6000mAh & 600 \\
     Companion Computer & Lattepanda & 55 \\
     Flight Controller & Pixhawk 6C & 16 \\
     Frame & Generative Designed Custom & 1340 \\\hline
    \end{tabular}
    \caption{Omnidirectional quadrotor components and weights. The total weight at take-off is $2.1kg$.}
    \vspace{-0.2cm}
\label{tab:drone_components}
\end{table}

\begin{figure}[t]
    \centering
    \begin{subfigure}[b]{0.49\linewidth}
        \includegraphics[width=\linewidth]{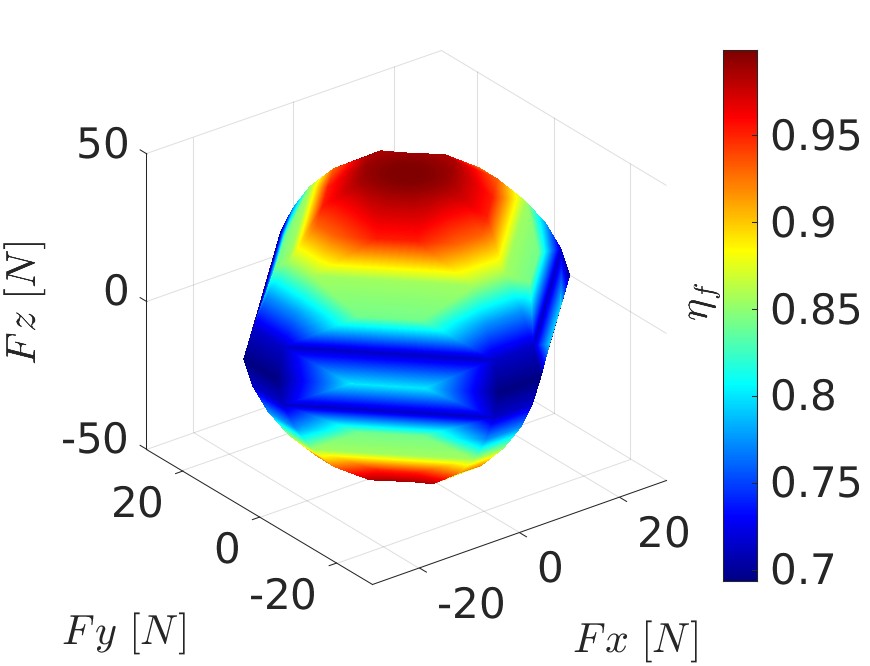}
        \caption{3D Force Envelope.}
        \label{fig:force_envelope}
    \end{subfigure}
    \hfill
    \begin{subfigure}[b]{0.49\linewidth}
        \includegraphics[width=\linewidth]{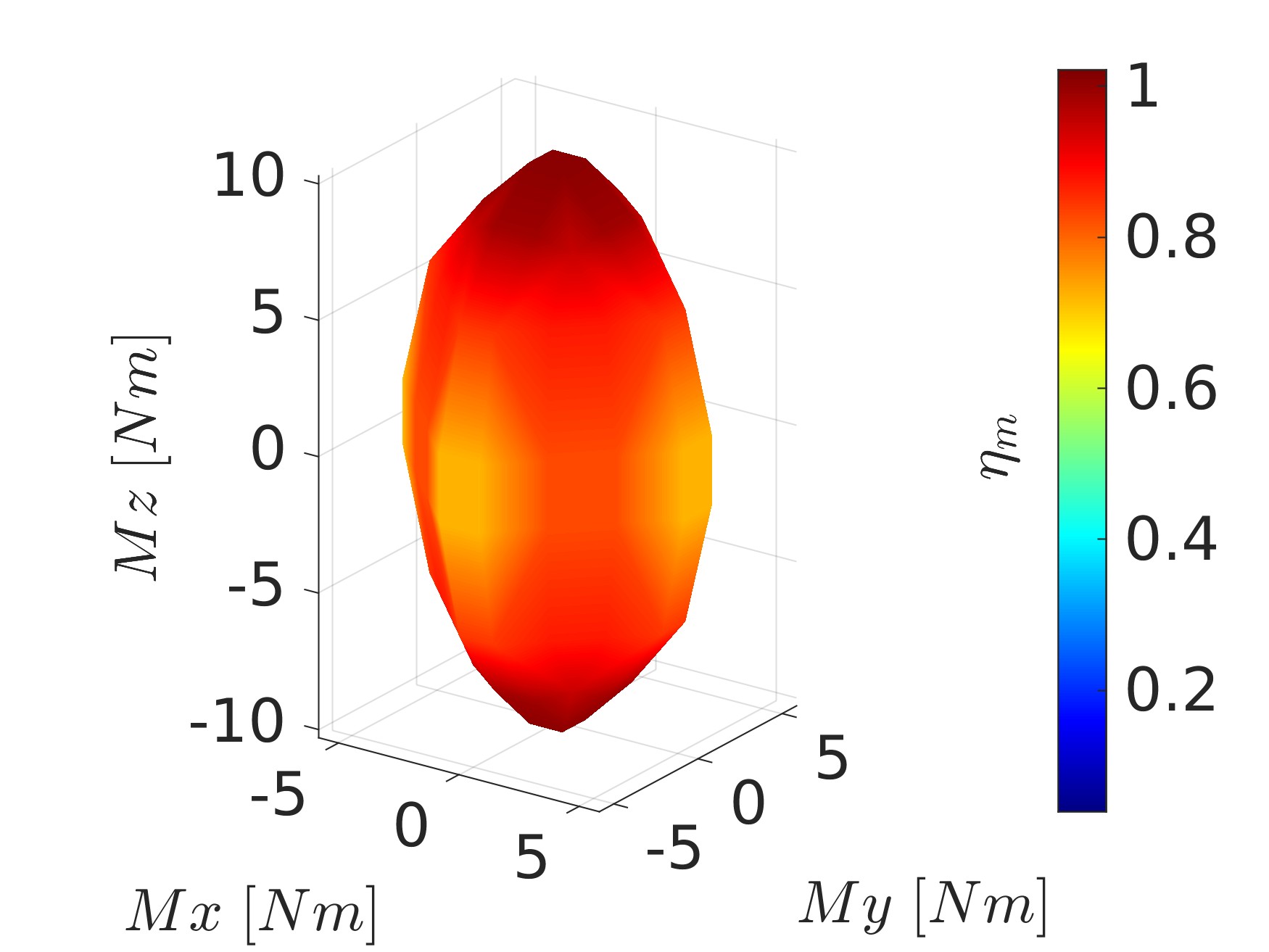}
        \caption{3D Torque Envelope}
        \label{fig:torque_envelope}
    \end{subfigure}
    \caption{3D force and torque representation with respectively thrust efficiency index $\eta_f$ and torque efficiency index $\eta_m$. Feasible values are achieved with servo angles $\alpha \in [-\pi, \pi]$ and rotor speeds within the maximum operational range, \mbox{$\omega \in [0, 2199.17]\text{rad/s}$}.}
    \label{fig:wrench_envelope}
    \vspace{-0.4cm}
\end{figure}

The frame design uses an optimization technique based on generative design methods~\cite{buonamici2020generative}. This approach involves specifying the internal wrench that occurs on the parts subject to mechanical stress. Generative design is advantageous as it facilitates the exploration of numerous design solutions by incorporating specific physical and force constraints~\cite{WANG2021101952}. With the proposed design, we then evaluate force and torque generation of the platform, by sending a set of forces and torque values to the mixer proposed in Sec.~\ref{sec:mixer}. The feasible values are represented by a force and torque set which are within the maximum servo angles of $\pi$ and maximum rotor speed of the adopted motor (Fig.~\ref{fig:wrench_envelope}]. From the results we see the platform capable of generating a maximum force of $50N$ on $\zB$ and $30N$ on $\xB$ and $\yB$ (Fig. \ref{fig:force_envelope}), and maximum torque of $5.8Nm$ on $\xB$ and $\yB$ and $10.3 Nm$ on $\zB$ (Fig. \ref{fig:torque_envelope}). We also evaluate the thrust efficiency $\eta_f$ and torque efficiency $\eta_m$ of the generated envelope defined in\cite{allenspach2020design}. They quantify the internal forces and torques dissipated in overactuated systems. In our case, the  $\eta_f \in [0.69,0.99]$, and $\eta_m \in [0.72, 1]$ demonstrating minimal internal losses and high efficiency.

\section{Haptic Device} 
\label{sec:haptics}
The proposed haptic device allows the operator to control the vehicle's velocity or attitude rate sending reference to the onboard flight controller. The inertial frame $\frameC=\{\originC,\xC,\yC,\zC\}$ with origin $\originC$ corresponds to the idle orientation of the control sticks with $\zR$ colinear to $\zC$ (see Fig.~\ref{fig:eye-catcher}). Their current poses are described by the frames $\frameL=\{\originL,\xL,\yL,\zL\}$ and $\frameR=\{\originR,\xR,\yR,\zR\}$ with respective origins $\originL$ and $\originR$ fixed with respect to the sticks. 

Attitude and angular rates of $\frameL$ and $\frameR$ with respect to $\frameC$ are defined as $\rotMatL, \rotMatR \in \SO{3}$, and $\angRotL, \angRotR \in \nR{2}$, respectively, the latter expressed in $\frameL$ and $\frameR$.

The dynamic relation between the haptic sticks values are modeled in $\frameL$ and $\frameR$ as
\begin{equation}
    \label{eq:operator-model}
    \mathbf{M}_{fing} \begin{bmatrix} \angVelL \\ \angVelR \end{bmatrix} + \mathbf{D}_{fing} \begin{bmatrix} \angRotL \\ \angRotR \end{bmatrix} = \begin{bmatrix} -\wrenchL \\ -\wrenchR \end{bmatrix} + \begin{bmatrix} \wrenchLAct \\ \wrenchRAct \end{bmatrix},
\end{equation}
where $\wrenchLAct, \wrenchRAct \in \nR{2}$ represents the torques generated by the muscles and $\wrenchL, \wrenchR \in \nR{2}$ represents the interaction torque with the haptic device. The inherent inertia and damping properties of the human operator are $\mathbf{M}_{fing} \in \nR{4 \times 4}$ and $\mathbf{D}_{fing} \in \nR{4 \times 4}$.

\subsection{Reference Generation}
As it has been proposed in~\cite{mike-6dof}, we use an admittance filter combined with a low-level joint position controller. This ensures compliant interaction with the fingers and haptic transparency. Assuming perfect tracking of the joysticks, the closed-loop dynamics can be approximated as
\begin{equation}
    \label{haptic_model_ref}
    \mathbf{M}_{adm} \begin{bmatrix} \angRotL \\ \angRotR \end{bmatrix} + \mathbf{D}_{adm} \begin{bmatrix} \angVelL \\ \angVelR \end{bmatrix} = \begin{bmatrix} -\wrenchL \\ -\wrenchR \end{bmatrix} + \wrenchTotal,
\end{equation}
where $\mathbf{M}_{adm} = diag(\mathbf{M}_{adm,l}, \mathbf{M}_{adm,r}) \in \nR{4 \times 4}$ is the inertia and $\mathbf{D}_{adm} = diag(\mathbf{D}_{adm,l}, \mathbf{D}_{adm,r}) \in \nR{4 \times 4}$ is the damping coefficient. Both are determined by the user's preferences. While $\wrenchTotal \in \nR{4}$ is the feedback wrench applied to the operator.

We establish the connection between the constrained input workspace and the limitless operational space of the robot, with velocity control. This is a well-recognized method in teleoperating aerial vehicles~\cite{finite-to-infinite}.
We define $v_1 , \omega_2 \in \nR{3}$ mapping orientations of the sticks $\rotMatCR, \rotMatCL \in \nR{3 \times 3}$ and orientation of the joystick body $\rotMatWC$ into linear and angular velocities
\begin{subequations}
    \label{eq:rot-input}
    \begin{IEEEeqnarray} {ll}
        v_1 &= \mathbf{P}_{1L} \mathbf{Q}(\rotMatL) + \mathbf{P}_{1R} \mathbf{Q}(\rotMatR) + \mathbf{P}_{1C} \mathbf{Q}(\rotMatC), \\
        \omega_2 &= \mathbf{P}_{2L} \mathbf{Q}(\rotMatL) + \mathbf{P}_{2R} \mathbf{Q}(\rotMatR) + \mathbf{P}_{2C} \mathbf{Q}(\rotMatC),
    \end{IEEEeqnarray}
\end{subequations}
with
\begin{equation}
    \mathbf{Q}(\rotMat) = \frac{(\rotMat - {\rotMat}^{\transpose})^{\vee}}{ \lVert (\rotMat - {\rotMat}^{\transpose})^{\vee} \rVert },
\end{equation}
and where $\mathbf{P}_{1L}, \mathbf{P}_{1R}, \mathbf{P}_{1C}, \mathbf{P}_{2L}, \mathbf{P}_{2R}, \mathbf{P}_{2C} \in \nR{3 \times 3}$ are a set of selection matrices given by the control mode to map the 6-\ac{DoF} inputs of the joysticks into the 6-\ac{DoF} capabilities of the omnidirectional platform chosen by the operator. To give outputs from the finite workspace of the stick to the potential infinite one of the robot, translational and rotational references are calculated such that
\begin{subequations}
    \label{eq:ref-input}
    \begin{IEEEeqnarray} {ll}
        \velRef &= \frac{\velMax}{2} \  v_1,\\
        \posRef &= \int_{0}^{t} \velRef(b) \,db,\\
        \Vec{\omega}_{B,\text{ref}}^B &= \frac{\omega_{max}}{2} \  \omega_2,\\
        \rotMatWBRef &= \int_{0}^{t} \rotMatWBRef(b) \left[ \Vec{\omega}_{B,ref}(b) \right]_{\times} \,db,
	\end{IEEEeqnarray}
\end{subequations}
where $(\cdot)_{\times} : \nR{3} \mapsto so(3)$ is the skew-symmetric operator, while $\velMax$ and $\omega_{max}$ are respectively the maximum velocity and angular rate set by the operator preference.

\subsection{Haptic Feedback Generation}
As proposed in~\cite{mike-6dof}, we define the total feedback torque
\begin{equation}
    \label{eq:total-wrench}
    \wrenchTotal = \wrenchRec + \wrenchFbExt,
\end{equation}
where $\wrenchRec \in \nR{4}$ is the recentering torque of each stick and $\wrenchFbExt \in \nR{4}$ is the interaction torque. In detail
\begin{equation}
    \label{eq:recentering-wrench}
    \wrenchRec = - \frac{\mathbf{K}_{rec}}{2} \begin{bmatrix} (\rotMatCL - {\rotMatCL}^{\transpose})^{\vee} \\ (\rotMatCR - {\rotMatCR}^{\transpose})^{\vee} \end{bmatrix},
\end{equation}
with $\mathbf{K}_{rec} = diag(\mathbf{K}_{rec,l} , \mathbf{K}_{rec,r}) \in \nR{4 \times 4}$ a tuning parameter. When the operator releases the stick, the recentering action causes it to return to the idle position, signifying a zero robot velocity. In other words, we replicate mechanical spring and with frictions of current \ac{RC} joysticks. An internal force-based impedance controller for each motor ensures modulation of force exertion. The torque feedback during interaction is defined as $\wrenchFbExt = \mathbf{K}_{ext} \wrenchEst$,
with \mbox{$\mathbf{K}_{ext} \in \nR{4}$} a tuning parameter to adjust perceived effort to user preferences. 

This modeling fits current teleoperation setups having conventional \ac{RC} and has the potential to extend the device to more \ac{DoF} per stick for overactuated systems. The formulation is independent of the device and allows the stick remapping of the four standard teleoperation control modes, known by \ac{MAV} pilots.

\subsection{Haptic Joystick Design}\label{sec:haptic_design}
For miniaturization purposes, the electronics are tailored and composed of two custom electronic boards. The first board connects the smart servos and the \ac{IMU} to the second board. This last embeds the powering with the microcontroller unit. Both designs are open-source and shared in the GitHub$^1$ repository. 

The proposed joystick embeds an IMU with a 3-DoF accelerometer and 3-DoF gyroscope. The attitude in quaternions is filtered out with the Madgwick filter~\cite{madgwick2010efficient}. After conversion to Euler angles, attitude around $\xC$ and $\yC$ with respect to $\frameW$ are part of the custom dataframe sent trough a serial link to the joystick driver. The other part of the dataframe has the 4-axis of the two joysticks describes by their respective orientations around $x$ and $y$ with respect to $\frameC$. Integration of these components into the joystick are illustrated in Fig.~\ref{fig:architecture_software} as part of the system.


At powering the device, a control routine on the microcontroller manages communications with multi-threading. Both centering and external torques defined in eq.~\ref{eq:total-wrench} are mapped and computed internally with an impedance controller. The servo motors can reach up to $0.441Nm$, or $22N$ force at the finger position on the sticks. This high force feedback capability ensures operator immersion with a wide range of force rendering.

\section{Experimental Applications}
\label{sec:experiment}

For this first application, the operator controls the aerial robot in translational velocities with the sticks in standardized for teleoperators $mode-2$ configuration. Also, orientation around $\xC$ and $\yC$ are mapped to the robot attitude in roll and pitch. In Eq.~\ref{eq:rot-input}, the matrices $\mathbf{P}_{1L}, \mathbf{P}_{1R}, \mathbf{P}_{1C}, \mathbf{P}_{2L}, \mathbf{P}_{2R}, \mathbf{P}_{2C}$ are sparse matrices where only following elements are equal to one: $ \{ (\mathbf{P}_{1L})_{3,2}, (\mathbf{P}_{2L})_{1,2},  (\mathbf{P}_{2L})_{2,1}, (\mathbf{P}_{1R})_{3,1}, (\mathbf{P}_{2C})_{1,2}, (\mathbf{P}_{2C})_{2,1} \}$, and zero otherwise.\\



\begin{figure}[t]
    \centering
    \includegraphics[width=\linewidth]{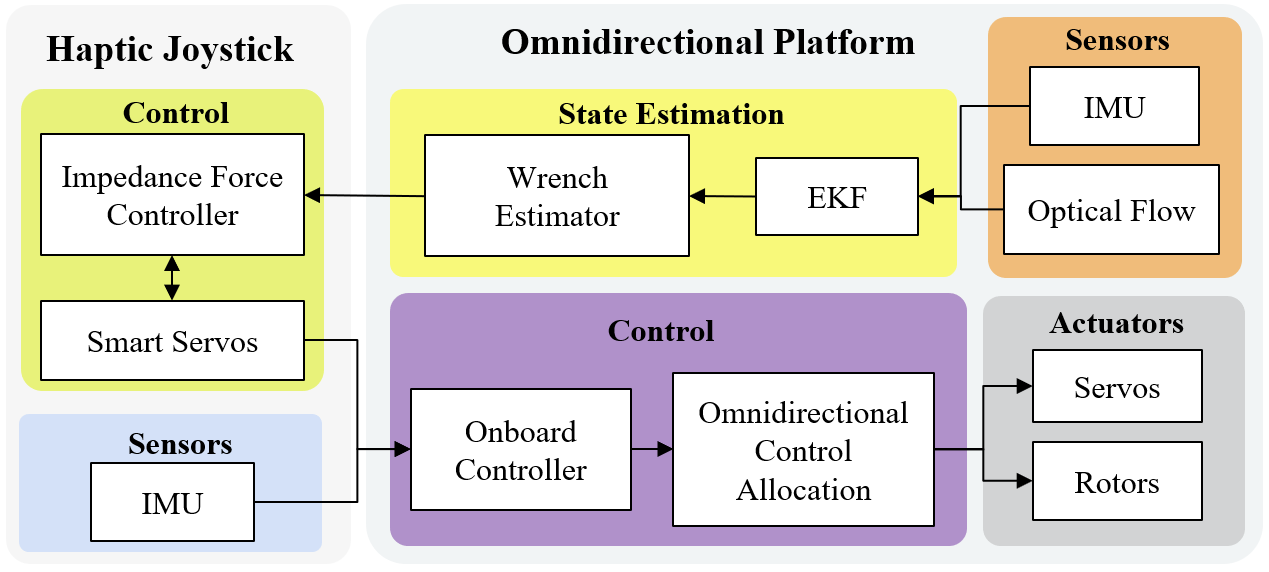}
    \caption{Interaction of telemanipulation system components.}
    \label{fig:architecture_software}
    \vspace{-0.4cm}
\end{figure}

The proposed telemanipulation system is composed of the haptic joystick and the omnidirectional robot. The platform presented in Sec.~\ref{sec:platform} uses PX4~\cite{marcellini2023px4} \ac{FC} both for the real platform and the \ac{SITL}. Modifications to integrate the controller and the haptic RC are shared online \footnote{https://github.com/tilties2/PX4-OmniQuad.git}. The haptic device detailed in Sec.~\ref{sec:haptics} links the platform through bilateral communication in ROS2. Fig. \ref{fig:architecture_software} illustrates the whole architecture.

\subsection{Single Axis Reference Generation} 
\begin{figure}[ht]
    \centering
    \includegraphics[width=1\linewidth]{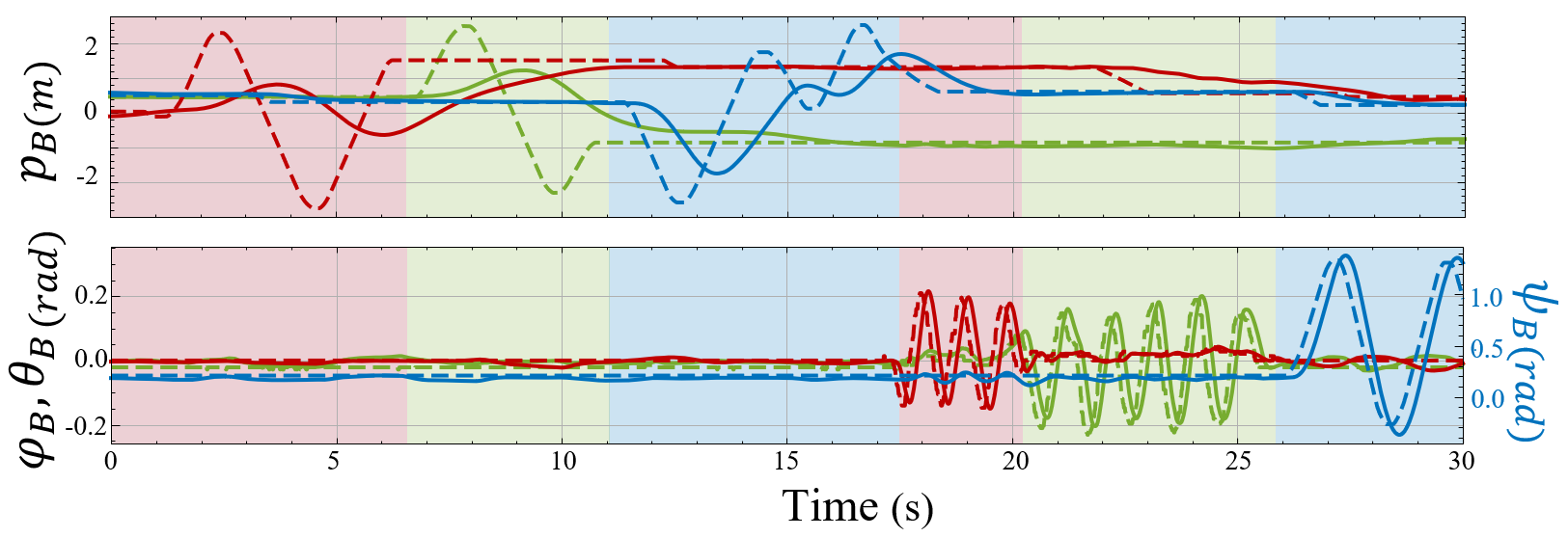}
    \caption{Illustration of single axis generation in dotted line, highlighted with shaded color. The robot state in plain line.}
    \label{fig:sitl_freeflight}
    \vspace{-0.4cm}
\end{figure}

Steering the vehicle in a single \ac{DoF} is an important property of the system. It simplifies control, enhancing precision for \ac{APhI} and maneuverability in complex industrial environments. To evaluate the reference generation from the real haptic device, it is appropriate to utilize \ac{SITL} within the physics engine simulator~\cite{ode-engine}. It connects all the components of the real hardware setup by simulating the physical environment and sensors with their characteristic noise. For each axis, we move back and forth during $30s$ in a single direction and evaluate the statistical influence on the other axis. A representation of the experimental plot is presented in Fig. \ref{fig:sitl_freeflight}, where no significant influence at moving one axis on another is observable. The extensive experiment is summarize in Table \ref{tab:single_axis_steer}. We compute the Mean Absolute Error (MAE) for each reference axis relative to every other axis.

\begin{table}[ht!]
\centering
\caption{MAE of repetitive axis command \textit{(column)} with respect to other axis \textit{(row)}.}\label{tab:single_axis_steer}
\scriptsize 
\begin{tabular}{ccccccc}
\toprule
&  $\dot{x}_B$ & $\dot{y}_B$ & $\dot{z}_B$ & $\dot{\phi}_B$ & $\dot{\theta}_B$ & $\dot{\psi}_B$  \\ 
\midrule
$\dot{x}_B$ & $\times$ & 0.04048 & 0.0580 & 0.00021 & 0.00023 &  0.0023 \\ 
\midrule
$\dot{y}_B$ & 0.258 & $\times$ & 0.023 & 0.0007 & 0.0021 & 0.0061 \\ 
\midrule
$\dot{z}_B$ &  0.0737 & 0.0302 & $\times$ & 0.00876 & 0.0213 & 0.00246 \\ 
\midrule
$\dot{\phi}_B$ & 0.0360 & 0.0417 & 0.2270 & $\times$ & 0.0599 & 0.0078 \\ 
\midrule
$\dot{\theta}_B$ & 0.0217 & 0.1010 & 0.0142 & 0.0126 & $\times$ & 0.1032 \\ 
\midrule
$\dot{\psi}_B$ & 0.0258 & 0.0596 & 0.1079 & 0.0020 & 0.0338 & $\times$ \\ 
\bottomrule
\end{tabular}
\vspace{-0.2cm}
\end{table}

Overall, the table indicates that the influence of one axis on the others is generally low, with MAE values remaining close to zero. While there are minor perturbations observed—such as on $\dot{x}_B$ when commanding $\dot{y}_B$ $(0.258)$ or on $\dot{z}_B$ when commanding $\dot{\phi}_B$ $(0.2270)$ these are non-significant. The data suggests that, the axes are largely decoupled, and the system exhibits good independent control with only minor perturbations between axes.

By keeping the conventional \ac{RC} form factor, our haptic device resolved the coupling of rotational reference with translational ones observed in~\cite{mike-6dof} and~\cite{rotation-ctl}. Explanation comes from the physical decoupling of each stick. Nevertheless, axes that are coupled in pairs have also little effect. Thus the proposed device should be considered for future aerial telemanipulation applications.

\begin{figure}[bt!]
    \centering
    \begin{subfigure}{\linewidth}
        \includegraphics[width=\linewidth]{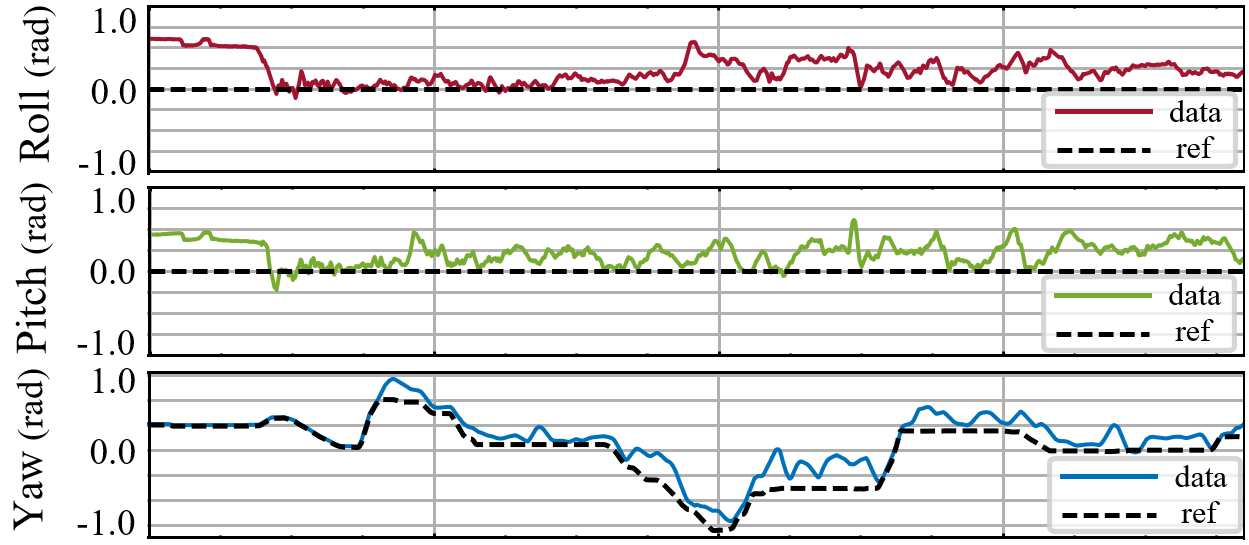}
        \caption{Attitude tracking performance with command reference.}
        \label{fig:track_flight}
    \end{subfigure}
    \hfill
    \begin{subfigure}{\linewidth}
        \includegraphics[width=\linewidth]{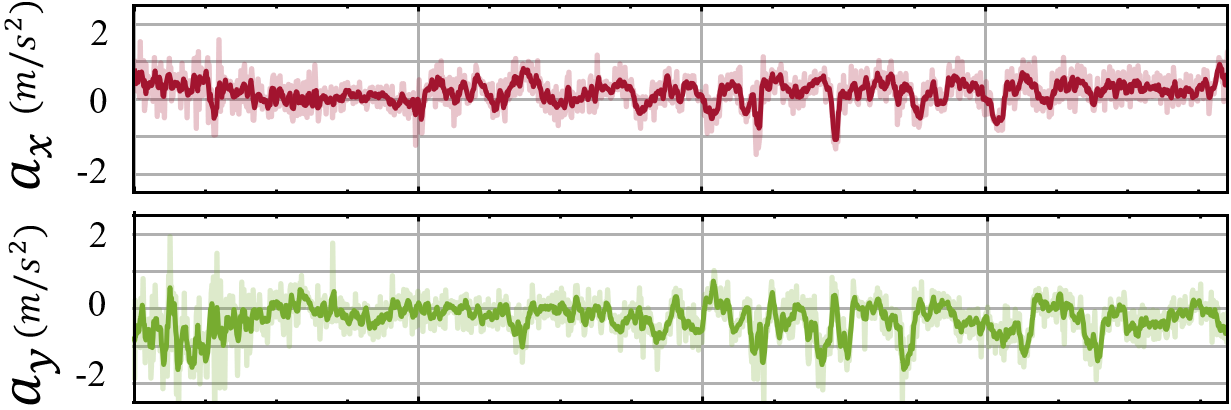}
        \caption{Measured acceleration from onboard inertial sensing.}
        \label{fig:acc_flight}
    \end{subfigure}
    \hfill
    \begin{subfigure}{\linewidth}
        \includegraphics[width=\linewidth]{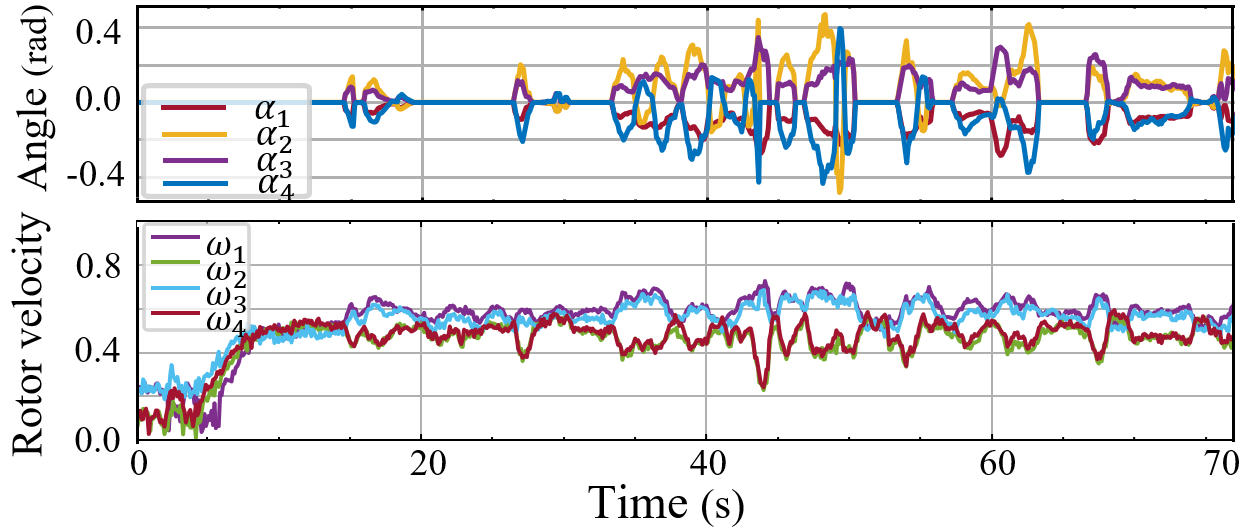}
        \caption{Servo angle state (top) and rotors normalized speed (bottom) during flight test.}
        \label{fig:servo_rotor_flight}
    \end{subfigure}
    \caption{Flight test result performances of the omnidirectional aerial robot in free flight.}
    \label{fig:rpy_track_flight}
    \vspace{-0.2cm}
\end{figure}

\subsection{Preliminary Flight Tests} 
To evaluate the flight capabilitis of the system, we perform two preliminary stable flights. In the first one, we make freeflight at horizontal attitude. In other words we want to command the platform is pure translations and yaw. As a result, the platform is capable of quickly translating in the plane $(\originB,\xW,\yW)$, as shown by its acceleration values (Fig.~\ref{fig:acc_flight}), while maintaining a horizontal orientation relative to the ground. Notably platform's measured roll and pitch (Fig.~\ref{fig:track_flight}) remains close to zero for the whole flight. The measured roll average angle is $0.026rad$ while pitch is $0.021rad$. Regards its actuation capability, the servo angles and rotor command are shown in Fig.~\ref{fig:servo_rotor_flight}. We observe continuous variations of the four servos angles to match the desired forces (Fig.~\ref{fig:servo_rotor_flight}). A slight difference between the rotor speed (Fig.~\ref{fig:servo_rotor_flight}) of the first and third rotor with respect to the second and fourth is noticeable at around $40sec$. It is the consequence of the decrease in the desired yaw (Fig.~\ref{fig:track_flight}). Then the rotor speeds consistently stay within $40\%$ to $60\%$ of their maximum value, with an average of $48.3\%$, demonstrating an efficient lift-to-mass ratio for the platform, as well as a good output of the proposed allocation matrix (Eq.~\ref{eq:damping_constant}).\\

The last experiment consists of going in contact and pushing against a vertical wall to show stable interaction and tilting capabilities of the aerial vehicle.

\begin{figure}[t]
    \centering
    \includegraphics[width=1\linewidth]{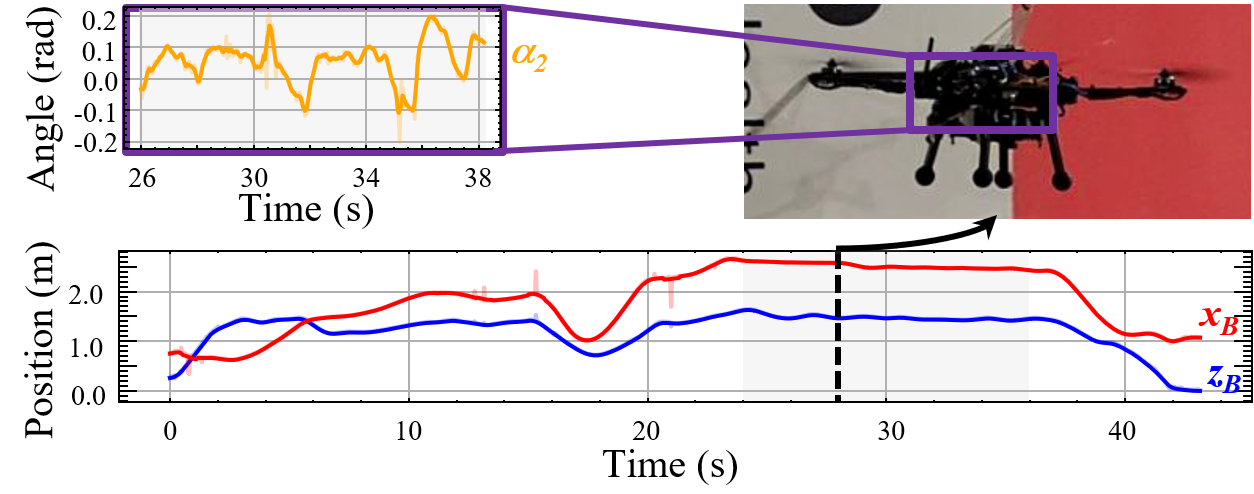}
    \caption{Plot of the robot navigation mission in the plane $(\originW,\xW,\zW)$ with a physical interaction shaded in grey. During the contact, we plot the rotor angle $\alpha_1$.}
    \label{fig:exp-contact}
    \vspace{-0.4cm}
\end{figure}

When the drone begins contact, vertical oscillations can be noticed in Fig.~\ref{fig:exp-contact} before stabilization at $t=30s$ by keeping pushing. During the $12s$ duration of physical interaction, $\alpha_1$ had an average of $0.05rad$ showing an overall push in the robot's forward direction. Some negative angles are still noticeable while the drone keeps contact. This can be explained by the overall internal force that is compensated by other rotors, bringing even more stability to the flying machine.

\section{Conclusion} 
\label{sec:conclusion}
By integrating a miniaturized haptic device into a conventional \ac{RC} shape, we resolved the problem of coupling rotation and translation motion of omnidirectional multirotor. The proposed platform allows us to reach state-of-the-art full actuation flight performance, while improving thrust-to-weight ratio and simplifying platform maintenance. All components of this system, both hardware and software, are open-sourced, making it accessible and modular for the wider research and development community. The presented system sets new opportunities for omnidirectional drone control, providing a highly efficient and user-friendly solution.
Future works will extend the generalized teleoperation framework to enable control and mapping of additional \ac{DoF}. In particular, we aim to provide the operator with control over the robot's internal forces. This would enhance platform stability, while we would question how to deliver feedback on the internal state through haptic modalities.

\bibliographystyle{IEEEtran}
\bibliography{references.bib}

\begin{thebibliography}{10}
\providecommand{\url}[1]{#1}
\csname url@samestyle\endcsname
\providecommand{\newblock}{\relax}
\providecommand{\bibinfo}[2]{#2}
\providecommand{\BIBentrySTDinterwordspacing}{\spaceskip=0pt\relax}
\providecommand{\BIBentryALTinterwordstretchfactor}{4}
\providecommand{\BIBentryALTinterwordspacing}{\spaceskip=\fontdimen2\font plus
\BIBentryALTinterwordstretchfactor\fontdimen3\font minus \fontdimen4\font\relax}
\providecommand{\BIBforeignlanguage}[2]{{%
\expandafter\ifx\csname l@#1\endcsname\relax
\typeout{** WARNING: IEEEtran.bst: No hyphenation pattern has been}%
\typeout{** loaded for the language `#1'. Using the pattern for}%
\typeout{** the default language instead.}%
\else
\language=\csname l@#1\endcsname
\fi
#2}}
\providecommand{\BIBdecl}{\relax}
\BIBdecl

\bibitem{past-future-am}
A.~Ollero, M.~Tognon, A.~Suarez, D.~Lee, and A.~Franchi, ``Past, present, and future of aerial robotic manipulators,'' \emph{IEEE Transactions on Robotics}, vol.~38, no.~1, pp. 626--645, 2022.

\bibitem{selvaggio2021haptics}
M.~Selvaggio, M.~Cognetti, S.~Nikolaidis, S.~Ivaldi, and B.~Siciliano, ``Autonomy in physical human-robot interaction: A brief survey,'' \emph{IEEE Robotics and Automation Letters}, vol.~6, no.~4, pp. 7989--7996, 2021.

\bibitem{bodie2020active}
K.~Bodie, M.~Brunner, M.~Pantic, S.~Walser, P.~Pf{\"a}ndler, U.~Angst, R.~Siegwart, and J.~Nieto, ``Active interaction force control for contact-based inspection with a fully actuated aerial vehicle,'' \emph{IEEE Transactions on Robotics}, vol.~37, no.~3, pp. 709--722, 2020.

\bibitem{botian2024omni}
B.~Xu, F.~Gao, C.~Yu, R.~Zhang, Y.~Wu, and Y.~Wang, ``Omnidrones: An efficient and flexible platform for reinforcement learning in drone control,'' \emph{IEEE Robotics and Automation Letters}, vol.~9, no.~3, pp. 2838--2844, 2024.

\bibitem{bilateral}
A.~Y. Mersha, S.~Stramigioli, and R.~Carloni, ``On bilateral teleoperation of aerial robots,'' \emph{IEEE Transactions on Robotics}, vol.~30, no.~1, pp. 258--274, 2014.

\bibitem{mike-6dof}
M.~Allenspach, N.~Lawrance, M.~Tognon, and R.~Siegwart, ``Towards 6dof bilateral teleoperation of an omnidirectional aerial vehicle for aerial physical interaction,'' in \emph{2022 International Conference on Robotics and Automation (ICRA)}, 2022, pp. 9302--9308.

\bibitem{rotation-ctl}
M.~Young, C.~Miller, Y.~Bi, W.~Chen, and B.~D. Argall, ``Formalized task characterization for human-robot autonomy allocation,'' in \emph{2019 International Conference on Robotics and Automation (ICRA)}, 2019, pp. 6044--6050.

\bibitem{mintchev2019portable}
S.~Mintchev, M.~Salerno, A.~Cherpillod, S.~Scaduto, and J.~Paik, ``A portable three-degrees-of-freedom force feedback origami robot for human--robot interactions,'' \emph{Nature Machine Intelligence}, vol.~1, no.~12, pp. 584--593, 2019.

\bibitem{brunner2022planning}
M.~Brunner, G.~Rizzi, M.~Studiger, R.~Siegwart, and M.~Tognon, ``A planning-and-control framework for aerial manipulation of articulated objects,'' \emph{IEEE Robotics and Automation Letters}, vol.~7, no.~4, pp. 10\,689--10\,696, 2022.

\bibitem{Kamel2018}
M.~Kamel, S.~Verling, O.~Elkhatib, C.~Sprecher, P.~Wulkop, Z.~Taylor, R.~Siegwart, and I.~Gilitschenski, ``{Voliro: An omnidirectional hexacopter with tiltable rotors},'' \emph{arXiv}, 2018.

\bibitem{hui2024passive}
\BIBentryALTinterwordspacing
T.~Hui, E.~Cuniato, M.~Pantic, M.~Tognon, M.~Fumagalli, and R.~Siegwart, ``Passive aligning physical interaction of fully-actuated aerial vehicles for pushing tasks,'' 2024. [Online]. Available: \url{https://arxiv.org/abs/2402.17434}
\BIBentrySTDinterwordspacing

\bibitem{buonamici2020generative}
F.~Buonamici, M.~Carfagni, R.~Furferi, Y.~Volpe, L.~Governi \emph{et~al.}, ``Generative design: an explorative study,'' \emph{Computer-Aided Design and Applications}, vol.~18, no.~1, pp. 144--155, 2020.

\bibitem{WANG2021101952}
\BIBentryALTinterwordspacing
Z.~Wang, Y.~Zhang, and A.~Bernard, ``A constructive solid geometry-based generative design method for additive manufacturing,'' \emph{Additive Manufacturing}, vol.~41, p. 101952, 2021. [Online]. Available: \url{https://www.sciencedirect.com/science/article/pii/S2214860421001172}
\BIBentrySTDinterwordspacing

\bibitem{allenspach2020design}
M.~Allenspach, K.~Bodie, M.~Brunner, L.~Rinsoz, Z.~Taylor, M.~Kamel, R.~Siegwart, and J.~Nieto, ``Design and optimal control of a tiltrotor micro-aerial vehicle for efficient omnidirectional flight,'' \emph{The International Journal of Robotics Research}, vol.~39, no. 10-11, pp. 1305--1325, 2020.

\bibitem{finite-to-infinite}
F.~Conti and O.~Khatib, ``Spanning large workspaces using small haptic devices,'' in \emph{First Joint Eurohaptics Conference and Symposium on Haptic Interfaces for Virtual Environment and Teleoperator Systems. World Haptics Conference}, 2005, pp. 183--188.

\bibitem{madgwick2010efficient}
S.~Madgwick \emph{et~al.}, ``An efficient orientation filter for inertial and inertial/magnetic sensor arrays,'' \emph{Report x-io and University of Bristol (UK)}, vol.~25, pp. 113--118, 2010.

\bibitem{marcellini2023px4}
S.~Marcellini, J.~Cacace, and V.~Lippiello, ``A px4 integrated framework for modeling and controlling multicopters with til table rotors,'' in \emph{2023 International Conference on Unmanned Aircraft Systems (ICUAS)}.\hskip 1em plus 0.5em minus 0.4em\relax IEEE, 2023, pp. 1089--1096.

\bibitem{ode-engine}
R.~Smith, ``{Open Dynamics Engine},'' 2008, http://www.ode.org/.

\end{thebibliography}

\end{document}